\documentclass[runningheads]{llncs}
\usepackage{graphicx}
\usepackage[sort]{cite}
\usepackage{booktabs}
\usepackage{multirow}
\usepackage{graphicx}
\usepackage{tabularx}
\usepackage{mathtools}
\usepackage{listings}

\usepackage[utf8x]{inputenc}
\usepackage{lmodern,textcomp}
\usepackage{threeparttable}
\usepackage[breaklinks=true,colorlinks,bookmarks=false]{hyperref}
\newtheorem{assumption}{Assumption}

\begin{document}
\title{TabEAno: \\Table to Knowledge Graph Entity Annotation}
\titlerunning{TabEAno}

\author{Phuc Nguyen\inst{1} \and
Natthawut Kertkeidkachorn\inst{2} \and \\
Ryutaro Ichise\inst{1,2} \and
Hideaki Takeda\inst{1}
}
\authorrunning{Phuc Nguyen et al.}
%
\institute{National Institute of Informatics, Japan \and National Institute of Advanced Industrial Science and Technology, Japan\\}
\maketitle 
\begin{abstract}
In the Open Data era, a large number of table resources have been made available on the Web and data portals. However, it is difficult to directly utilize such data due to the ambiguity of entities, name variations, heterogeneous schema, missing, or incomplete metadata. To address these issues, we propose a novel approach, namely TabEAno, to semantically annotate table rows toward knowledge graph entities. Specifically, we introduce a ``two-cells” lookup strategy bases on the assumption that there is an existing logical relation occurring in the knowledge graph between the two closed cells in the same row of the table. Despite the simplicity of the approach, TabEAno outperforms the state of the art approaches in the two standard datasets e.g, T2D, Limaye with, and in the large-scale Wikipedia tables dataset.
\keywords{Entity linking  \and Tabular data \and Table annotation \and Knowledge Graph}
\end{abstract}

\section{Introduction}
Thanks to the Open Data community\footnote{Open Data Vision: \url{https://opendatabarometer.org}}, a large number of table resources have been made available on the Web and data portals. As a study of Lehmberg et al. \cite{Lehmberg2016}, there are 233 million table resources extracted from Common Crawl July 2015\footnote{Common Crawl: \url{http://commoncrawl.org/}}. Mitlohner et al. also analyze about 200,000 table resources from 232 Open Data Portals \cite{Mitlohner2016}. However, the usage of such tables is limited due to the ambiguity of entities, name variations, heterogeneous schema, missing, or incomplete metadata. 

To improve the usability of these table data, we need to resolve the meaning of elements in a table by annotating table elements into knowledge graph concepts. The annotated tables are potentially useful for a number of applications, such as table search \cite{Nguyen2015, Nargesian2018}, table extension \cite{Lehmberg2015}, completion\cite{ahmadov2015towards}, or knowledge graph construction as used in DBpedia \cite{Zhang2013}, YAGO \cite{Sekhavat2014}, and Freebase \cite{Dong2014}. 

In this paper, we focus on the table to the knowledge graph entity annotation problem. Note that, tables in this study refer to relational vertical tables. A relational vertical table is a table that contains the core column (the core table attribute) and the attributes columns. In the core column,  target entities are listed, while the other columns are the attributes using to describe each entity in the core table attribute. Typically, this problem also known as record linkage \cite{hassanzadeh2012record}, or data, or instance-level matching \cite{ritze2015matching}. In particular, table rows are matched into knowledge graph entities. In this study, DBpedia is used as the target knowledge graph for table rows annotation. 

The most common pipeline of table entity annotation has three steps:  1) structure annotation, 2) candidate generation, and 3) candidate disambiguation. Most prior works focus on the candidate disambiguation step where the best entity candidate is determined in order to achieve the best relevance score of the table row. However, TabEAno targets on improving the structure annotation and the candidate generation steps, whereas we adopt the simple aggregation to re-rank entity candidates in the candidate disambiguation step.  Specifically, we improve the heuristic of the core attribute and header annotation in the structure annotation step. Regarding the candidate generation step, we propose a “two-cells” lookup strategy. In the strategy, we assume that there is an existing logical relation occurring in the knowledge graph between the two closed cells in the same row of the table. Despite the simplicity of the approach, TabEAno outperforms the other baselines approaches in terms of annotation ability in the two standard datasets e.g, T2D, Limaye, and the large-scale Wikipedia tables dataset.

To sum up, the contributions of our paper are as follows:
\begin{itemize}
    \item We propose new heuristic rules for core attribute annotation, and header annotation in Section \ref{subsec:step1}. These heuristic rules improve the accuracy performance of the core attribute annotation 3\% and 1\% in the on T2D dataset and the Wikipedia tables dataset respectively.  
    \item We introduce a novel lookup strategy, namely ``two-cell" lookup for entity candidate generation step in Section \ref{subsec:step3}. 
    \item We empirically study and evaluate TabEAno on the two standard datasets e.g, T2D, Limaye, and the large-scale Wikipedia tables dataset\cite{efthymiou2017matching} which contains about 480 million extracted tables from Wikipedia pages
    \item Also, the repository of the TabEAno system will be made available online to further facilitate the research in this field.
\end{itemize}

The rest of this paper is organized as follows. In Section \ref{sec:relate_work}, we discuss the related works on the task of table entity annotation for table data, and prior approaches on table annotation approaches, and the difference between TabEAno and other prior works. In Section \ref{sec:assumptions}, we present the assumptions of TabEAno. Then, we present the TabEAno in Section \ref{sec:approach}. In Section \ref{sec:evaluation}, we describe the details of our evaluation, and experimental settings then present the results. Finally, we summarize the paper and discuss the future direction in Section \ref{sec:conclusion}.

\section{Related Work} \label{sec:relate_work}
In this section, we summaries the related work on table annotation as well as the differences between TabEAno and other approaches. 
\subsection{Annotation Tasks}
We categorize the tasks of table annotation as two main tasks e.g., structure, and semantic annotation. 

The structure annotation contains many sub-tasks as table type prediction\cite{Nishida2017}, data type prediction, table header annotation, core attribute prediction, and holistic matching across tables \cite{Lehmberg:2017:SWT:3137628.3137657}. In this paper, we assume that the input tables are relational vertical types, and these are independence to each other. So that, we only perform the sub-tasks of data-type prediction, table header annotation, and core attribute prediction in TabEAno. 

Semantic annotation involves of matching table elements into knowledge concepts such as schema-level matching e.g, tables to classes\cite{ritze2015matching}, columns to properties\cite{ritze2015matching,pham16:iswc,colnet, embnumplus} or classes\cite{zhang2017effective}, and data-level matching e.g., rows \cite{ritze2015matching, efthymiou2017matching} or cells \cite{DBLP:journals/pvldb/LimayeSC10, zhang2017effective, DBLP:conf/semweb/2019semtab} to entities. In this study, we focus on the task of rows to entities annotation. 

\subsection{Supervised and Unsupervised Setting}
The most popular approaches to row-to-entities annotation as a supervised learning setting, where entities candidates are selected by a classification model learned from a subset of the gold standard \cite{DBLP:journals/pvldb/LimayeSC10, ritze2015matching, tabel}. However, the gold standard is not completed and perfect so that the model learned from one dataset might not work well in different or a large-scale dataset. In this study, we focus on the unsupervised setting as \cite{efthymiou2017matching} which does not make any assumption about the availability of training data. 

\subsection{Candidate Generation Approaches}
Regarding finding relevance entity candidates, Ritze et al. \cite{ritze2015matching} use the DBpedia lookup service \footnote{DBpedia Lookup: \url{https://wiki.dbpedia.org/Lookup}} which is only index DBpedia dump knowledge. Due to the situation that DBpedia has many interlinked resources, we can use other lookup services and redirect to the DBpedia IRI entity by there mappings such as Freebase lookup \cite{zhang2017effective}, Wikipedia lookup \cite{www2020}. Also, combining multiple resources e.g., Wikipedia, Wikidata, Freebase, YAGO proved effective as FactBase\cite{efthymiou2017matching}, and MTab \cite{mtab}. Different from multiple resources aggregation, we only use the DBpedia dump as indexing in TabEAno. It makes TabEAno more practical when applying the method to a different knowledge graph when we do not have interlinked resources to the target knowledge graph. Additionally, the lookup results of TabEAno are easy to re-produce with the dump data of DBpedia.

\subsection{Candidate Disambiguation Approaches}
Regarding candidate disambiguation, there are many useful context features were considered as the prior works such as textual information e.g., entity candidate types (including the direct type, and transitive types), entity labels, entity attributes \cite{DBLP:journals/pvldb/LimayeSC10, ritze2015matching, zhang2017effective, efthymiou2017matching, www2020}, semantic similarity using pre-trained word embeddings \cite{efthymiou2017matching, www2020}, or numerical information as distribution of numerical columns \cite{mtab}. However, in this study, we only adopt the five features derived directly from the target knowledge graph without taking the effort of pre-training or prior knowledge. These five features are two-cell lookup results, direct types, transitive types, surface similarities of entity labels, and the entities attributes with value matching. 

\section{Assumptions} \label{sec:assumptions}
In TabEAno, we adopt the following assumptions:
\begin{assumption} \label{closed-world}
TabEAno is built on a closed-world assumption.  
\end{assumption}
It means that the target knowledge graph (DBpedia) is completed and corrected. The table rows could only be matched when there is a corresponding entities in knowledge base.
\begin{assumption} \label{Table_types}
The type of a input table is the relational vertical type.
\end{assumption}
 In other words, the input table have one column is the table core attribute that contains mentions about entities. The other columns represent the relation between the entities and their attribute values. 
\begin{assumption} \label{Independece}
Input tables are independence, and there is no assumption about data sharing across input tables. 
\end{assumption}
\begin{assumption} \label{Column}
Column cell values have the same data types and carry a same semantic type or context. 
\end{assumption}
\begin{assumption} \label{Header}
Table headers are located in some of the first rows of a table, and there is a difference between the length of header values and the length of data values. 
\end{assumption}

\section{TabEAno Approach} \label{sec:approach}
\begin{figure}[!ht]
\centering
\includegraphics[width=1\textwidth]{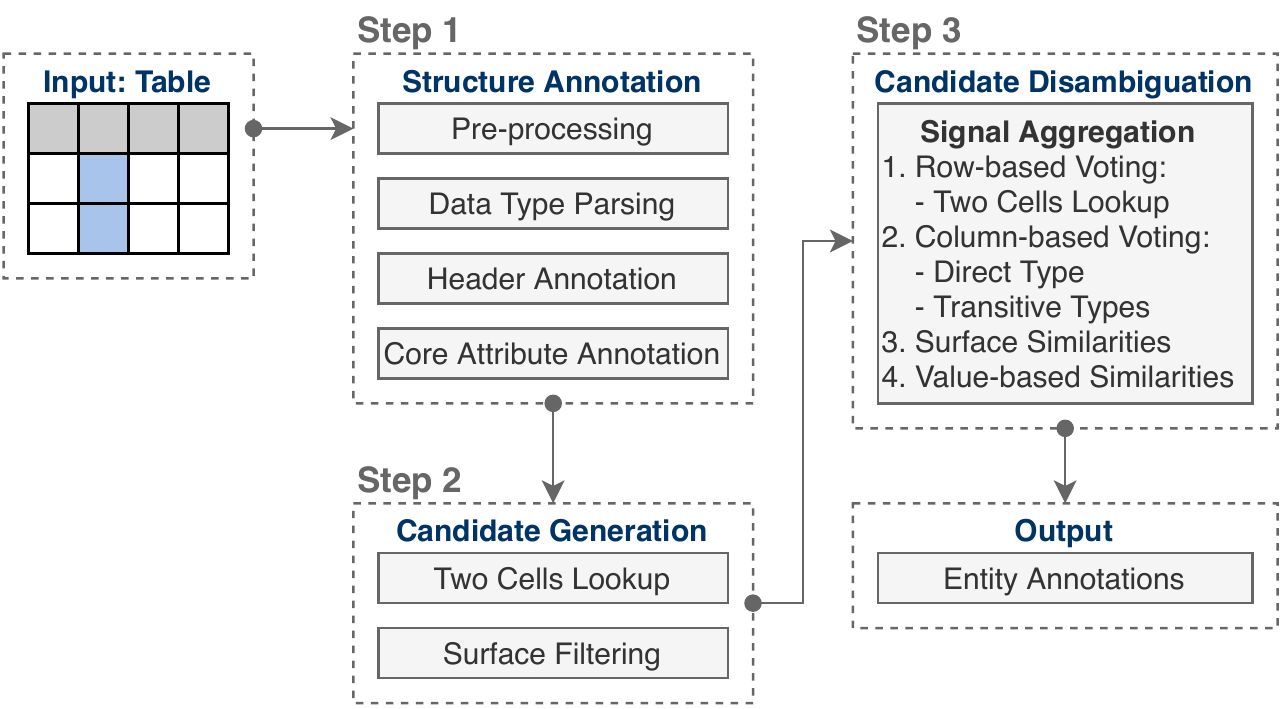}
\caption{The overview of the TabEAno framework} 
\label{fig:framework}
\end{figure}

In this section, we present the overall framework of TabEAno for the task of entity annotation for table rows as in Figure \ref{fig:framework}. Due to the problem of gold standards completeness and correctness, we design TabEAno to work as an unsupervised setting, and it does not rely on any training data.

Given a knowledge graph and an input relational vertical table, we perform table row annotation toward knowledge graph entities. In particular, we first perform structure annotations e.g., data type prediction, header annotation, and core attribute annotation (Section \ref{subsec:step1}), and then, we perform the candidate generation with ``two cells" lookup and surface filtering in Section \ref{subsec:step2}. Finally, the entity annotations are derived from the candidate disambiguation in Section \ref{subsec:step3} by aggregating the five features e.g. lookup results, direct types, transitive types, surface similarities, and values matching. 

\subsection{Step 1: Structure Annotation} \label{subsec:step1}
\subsubsection{Pre-Processing} \label{sssec:preprocessing}
As the initial step, due to the problem of text encoding in the Open Data we perform table cell cleaning. This step is essential to alleviate the error in the following steps. We perform several steps as follows.
\begin{itemize}
    \item Fix text decoding: Table cells might contain many noisy sequences as the problem of incorrect text decoding, e.g. an em dash ``—" might appear as ``â€". We use the ftfy tool \cite{speer-2019-ftfy} to fix all noisy or incorrect decoding textual data in tables.
    \item  As the input data could be extracted from the Web so that it contains many noisy HTML tags. We remove noisy characters such as HTML tags, e.g. ``\&nbsp", ``\&amp;", and punctuation.
    \item We also normalize accented characters of table cells into ASCII character form, e.g. ``café” is normalized to ``cafe”.
\end{itemize}

\subsubsection{Data Type Parsing} \label{sssec:data_type_parsing}
In this step, we perform data type prediction for each table cells whereas these are numerical values or textual values. We use Duckling\footnote{Duckling,  \url{https://github.com/facebook/duckling}} and SpaCy Named Entity Recognition (NER)\cite{spacy2} to parse data types of table cells. The final types of cell values are based on the majority vote of the parsing results of Duckling and SpaCy. 

In particular, Duckling is a tool to parse text to 13 data types based on Probabilistic Context-Free Grammar. Those types are about numerical tags, e.g., amount of money, credit card number, distance, duration, number, ordinal,  phone number, quantity, temperature, time, volume, or special tags, e.g., email, URL. If there is a data type recognized, we assigned it as a numerical cell. If there is no data type assigned, we assign this cell type as a textual cell. In SpaCy NER, the entity types are derived from pre-trained deep learning model. We use the pre-trained SpaCy NER model on OntoNotes 5 dataset to parse cell values to 18 entity types. We also map these 18 entity types to two consider types as textual (11 types, e.g. PERSON, ORG, LOC, NORP, FAC, GPE, PRODUCT, EVENT, WORK\_OF\_ART, LAW, and LANGUAGE) and numerical cells (7 types, e.g. CARDINAL, PERCENT, MONEY, ORDINAL, QUANTITY, TIME, and DATE). 
\subsubsection{Header Annotation} \label{ssec:header}
To annotate table headers, we use simple heuristics for header recognition follows:
\begin{itemize}
    \item We consider the first row of a table as the header candidate. 
    \item If the list of data types of the header candidate row differ from the list of the majority data types of the remaining rows, the candidate is the table header. For example, the list of data types of header candidate is [text, text, text], while the list of the majority data type of remaining rows are [text, number, number].
    \item We also found that the length of header text is usually sorter or longer the remaining rows. So that, we define that if the length of values of the header candidate row is less than the 0.05 quantile or larger than the 0.95 quantile of the length of the value of remaining rows, the candidates are the table header. 
\end{itemize}
\subsubsection{Core Attribute Annotation} \label{sssec:core_attribute}
To predict the core attribute of a table, we adopt the heuristics proposed by Ritze et. al. \cite{ritze2015matching} as well as modify a simple heuristic for the core attribute column as follows.
\begin{itemize}
    \item A column is a core attribute when its data type is textual. The data type of a column is based on the majority voting of the cells in this column. 
    \item The average cell values length is from 3.5 to 200. In this step, we add a restriction that only consider non-header cells since the length of table headers could very differ from the remaining cells. 
    \item The core attribute is determined based on the uniqueness score as an increased score for those columns with many unique values and reduce the score for those columns with many missing values. The core attribute is the highest unique score columns. If we have many columns that have the same score, the left-most column is chosen. 
\end{itemize}
In sum, the differences between our system with \cite{ritze2015matching} are that we use Duckling and SpaCy as data type annotation while Ritze et. al. \cite{ritze2015matching} manually defined 100 regular expressions. Additionally, we modified the heuristic rule to make the exception for the table headers in the process of annotating the table core attribute. 

\subsection{Step 2: Candidate Generation} \label{subsec:step2}
The candidate generation is the process of finding the relevant entity in a knowledge graph. We can find the correspondences by using knowledge from the target knowledge graph such as DBpedia Lookup \cite{ritze2015matching}. Other studies use the other resources which are interlinked with DBpedia such as Freebase Lookup \cite{zhang2017effective}, and Wikipedia Lookup \cite{www2020}. Aggregating multiple data resources e.g., Wikipedia, Wikidata, Freebase, YAGO also proved effective as FactBase \cite{efthymiou2017matching}, and MTab \cite{mtab}. 

However, due to the rapid change of knowledge graphs, the number of entities as well as entities content, and entity popularity in a knowledge graph differ across released knowledge graph versions. Moreover, using the interlinked lookup results from other resources also have many inconsistencies. As a result, it is difficult to reproduce the same lookup result if the tested knowledge graph and other resources are not the same time frame and consistence. 

Due to this difficulty, in this study, we only use the knowledge from DBpedia dump with indexing by Virtuoso\footnote{Virtuoso: \url{http://vos.openlinksw.com/owiki/wiki/VOS}}. We propose a novel entity search as two steps. First, we perform a ``two-cells" lookup based on SPARQL facet queries. Second, we only select those entity candidates that have the most relevance in terms of string similarity. The two steps are described as follows.

\begin{figure} [!ht]
\centering
\includegraphics[width=0.9\textwidth]{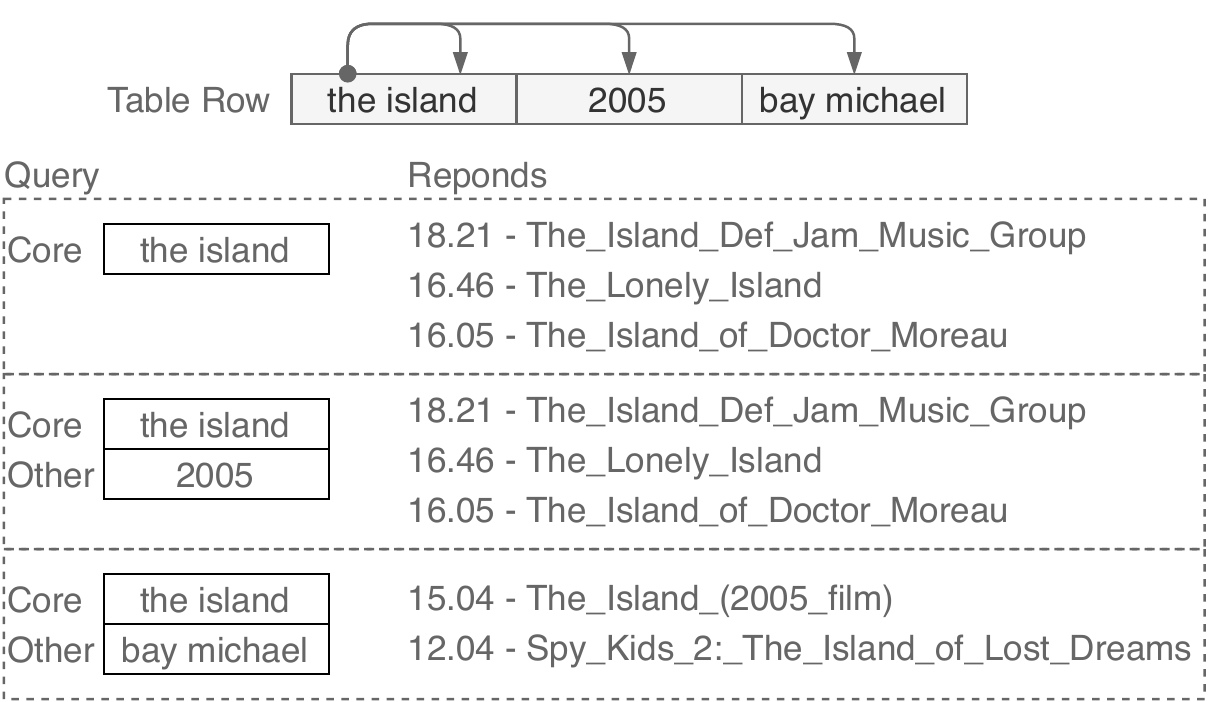}
\caption{Example of ``two cells" lookup for a row of the table of 40844462\_1\_6230938203735169234 in T2D dataset. The cell of ``the insland" is the core cell, while ``2005" and ``bay michael" are the other cells.} 
\label{fig:lk}
\end{figure}

\subsubsection{Two-cell Lookup} \label{sssec:lk}
TabEAno assumes that there is an existing logical relation occurring in the knowledge graph between the two closed cells of the table. Therefore, we introduce a simple ``two cells" lookup based on facet SPARQL search. In general, we can perform n-cells queries but we only consider ``two cells" to avoid too many combinations of table cells and preserve the scalability of the approach. The first cell of the ``two cells" is the core cell in the core attributes where the second cell is another cell in the same row with the core cell. 

Figure \ref{fig:lk} depicts an example of a two-cell lookup for a row in the table of 40844462\_1\_6230938203735169234 in T2D dataset. The core cell is ``the island", and the other cells are ``2005" and ``bay michael". We perform the one cell lookup for the core cell, then two query lookup for the two remaining cells as in Query \ref{query1}. The results are a list of relevance entity IRI ordered by the facet ranking\footnote{This feature is one of the Virtuoso packages. Note: The online DBpedia SPARQL \url{https://dbpedia.org/sparql} does not have entity ranking. To use these features, we need to install DBpedia locally with Virtuoso from OpenLink Software and run pre-compute text indexing and entity ranking}. The facet ranking ``?r" is the combination of text hit score and entity rank \cite{DBLP:conf/www/ErlingM09}. The entity rank is derived from inbound and outbound links of this entity.  

\begin{lstlisting}[caption=An example of the two-cell lookup for the cells of ``the island" and ``bay michael", label=query1,basicstyle=\ttfamily,frame=single]
SELECT DISTINCT ?s1 (?sc * 3e-1 + sql:rnk_scale 
                    (<LONG::IRI_RANK> (?s1))) as ?r
{
 ?s1 rdfs:label ?l1.
 ?l1 bif:contains `THE AND INSLAND' OPTION (score ?sc).
 ?s1 ?p1 ?s2.
 ?s2 ?p2 ?l2.
 ?label2 bif:contains `BAY AND MICHAEL'.
 FILTER regex(str(?s1), `http://dbpedia.org/resource/')
 FILTER !regex(str(?s1), `_(disambiguation)')
}
ORDER BY DESC (?sc * 3e-1 + 
               sql:rnk_scale (<LONG::IRI_RANK> (?s1)))
\end{lstlisting}

\subsubsection{Surface Filtering} \label{sssec:surface}
We further use the edit distance (Levenshtein distance) to measure the relevance in terms of lexical between the entity candidate labels and the core cell. Because of the efficiency reason, we only select the top three relevance entity candidates for each query. 

In our experiments, we use the average of the four similarities from the FuzzyWuzzy package\footnote{FuzzyWuzzy \url{https://github.com/seatgeek/fuzzywuzzy}} e.g. ratio, partial ratio, token sort ratio, and token set ratio. The ratio score is based on the Levenshtein distance of the total number of elements in two strings. This similarity could address the spelling mistakes of compared strings. The partial ratio is calculated based on the best partial matching with the shorter string. The token sort ratio put more relax for token order, while the token set ratio relaxes on the duplicate token in the compared strings. 

\subsection{Step 3: Candidate Disambiguation} \label{subsec:step3}
To select a best entity annotation from those entity candidates, we take the average score of the five signals from the results of ``two cell" lookup, direct types, transitive types, surface similarities, value matching. Then, we select the entity candidate with the highest average score as the final annotation.  

\subsubsection{Two-Cells Lookup Results}
We transform the ranking score of lookup results into entity probabilities using the softmax function. Then, take the average of entities candidates for all ``two cell" queries in the same rows. Overall, we have a probability of entity candidates given ``two cells" lookup for each row in tables. 
\subsubsection{Direct Types}
We take the aggregate of the probability of entity candidates in the previous step with their direct types as the score of direct types for the core columns. Then we transform the aggregated score to the probability of direct types using the softmax function and associated this as the probability of entity given direct types of columns. 
\subsubsection{Transitive types}
To calculate the probability of entity given transitive types of columns, we also do the same process as direct types calculation, however, instead of aggregate the direct type, we aggregate all the transitive types of entities. Noted that, in this step, we omit the type which is too general as Thing and Agent as the setting in SemTab 2019 \cite{DBLP:conf/semweb/2019semtab}. 

\subsubsection{Surface Similarities}
In this step, we use the same aggregated similarities in Section \ref{sssec:surface} to get similarities between entity candidates and the core cell value. 

\subsubsection{Value Matching}
In this step, we perform similarities between other cell values with the values of mapping objects, and literal values, as well as the infobox properties of each entity. We use the same similarities as in Section \ref{sssec:surface}. Only entities have value similarities larger than a value matching threshold is considered. 

\section{Evaluation} \label{sec:evaluation}
\subsection{Dataset} \label{ssec:dataset}
We use three gold standards datasets, e.g, T2D\cite{ritze2015matching}, Limaye, and Wikipedia tables \cite{efthymiou2017matching}. The detailed statistic of these datasets is reported in Table \ref{tab:dataset}. To verify whereas an entity IRI available in the knowledge graph or not, we use the ASK SPARQL to check if the IRI in the subject, predicate, subject of RDF triple in this knowledge graph. 

\begin{table}[!ht]
\centering
\caption{Statistic of the gold standards}
\label{tab:dataset}
\setlength{\tabcolsep}{0.8em} 
{\renewcommand{\arraystretch}{1.3}
\begin{tabular}{l|lll|ll}
\hline
\textbf{Dataset}   & \textbf{Tables}  & \textbf{Rows}      & \textbf{Matches}   & \textbf{DBpedia} & \textbf{Invalid IRI} \\ \hline
\textbf{T2D}       & 233     & 28,647    & 26,124    & 2014    & 27          \\
\textbf{Limaye}    & 296     & 8,670     & 5,278     & 2015-10 & 48          \\
\textbf{Wikipedia} & 485,096 & 7,437,606 & 4,453,329 & 2015-10 & 33,075     \\ \hline
\end{tabular}}
\end{table}

T2D (version 1) consists of 233 tables and 26,124 samples of row matching into DBpedia entities version 2014. There are 27 samples which are not available in the target Knowledge Graph. The dataset is well annotated with 91.19\% matched correspondences, while the 8\% remaining is non-matched samples. 

Limaye and Wikipedia tables datasets are both extracted from Wikipedia. Limaye dataset was adapted to map table rows into DBpedia entities version October 2015. The original Limaye dataset annotates table cells with Wikipedia pages \cite{tabel}. The two datasets have less ratio of matched correspondences (about 60\%) than the T2D dataset. The ratio of non-matched correspondences is about 40\%. 

As our analysis, the Wikipedia tables are not well annotated for non-matched correspondences since many table rows could be matched to correspondences but it is not available in the gold standard. Moreover, regarding the matched correspondences, we also found 767 disambiguation IRI samples in the Wikipedia tables dataset. These disambiguation correspondences are not a good annotation since it does not provide explicit meaning as an annotation.

\subsection{Evaluation Metrics} \label{ssec:metric}
To evaluate the annotation performance, we use similar metrics as micro-averaged precision, recall, and F1-score with the other baseline works. However, we want to emphasize the importance of the recall metrics to measure the ability to make more annotations. As the problem of non-matched correspondences in the Wikipedia tables dataset, the more annotations are made, the better quality of the annotation model is. 

\subsection{Setting and Implementation Details} \label{ssec:setting}
In our experiments, we use Virtuoso ver 07.20.3230 of OpenLink Software to index the English version of the DBpedia dump version of 2014\footnote{DBpedia 2014: \url{http://data.dws.informatik.uni-mannheim.de/dbpedia/2014/}}, and version of October 2015\footnote{DBpedia 2015-10: \url{http://downloads.dbpedia.org/2015-10/core/}}. Then, we perform pre-compute of text ranking (around 8 hours), and entity ranking (around 8 hours) for each DBpedia dump version. To make a fair comparison with baselines, we conduct TabEAno on T2D dataset using the DBpedia version 2014, and Limaye, Wikipedia tables dataset using DBpedia version October 2015. 

We set the parameter of a minimum of text similarity of surface filtering is 0.8, the minimum threshold of text similarity in Value Matching is 0.9. We consider the first row of the table as the header candidate in header annotation. The maximum number of candidates of entity candidate generation step is 3.

We perform a ``pre-two cell" lookup with multi-processing (36 sub-processes). It takes 1 second for 20 queries. The largest dataset is the Wikipedia tables with around 15 million queries takes around 9 days to obtain all lookup results. The entity disambiguation step takes several minutes for T2D, and Limaye dataset, while it takes a few hours in the Wikipedia tables dataset.

\subsection{Experimental Results}
In this section, we report the experimental results of the core attribute annotation and entity annotation. Note that, we do not report the results of the header annotation because the gold standard does not contain the ground truth for the table headers. 

\subsubsection{Core Attribute Annotation}
The core attribute annotation is an important task since it strongly affects the remaining module performance. However, not so many studies reported the results in this task. Most studies adopt a simple heuristic that the left-most textual column of a table is the core attributes, while the table uniqueness is considered in Ritze et al. \cite{ritze2015matching}.

In Table \ref{tab:core_att}, we report the results of the core attribute annotation of TabEAno and T2K in terms of accuracy metric. The performance of TabEAno achieves higher performance T2K with a simple modification as Section \ref{ssec:header}. 
\begin{table}[!ht]
\centering
\caption{Table Core Attribute Annotation Results in Accuracy}
\label{tab:core_att}
\setlength{\tabcolsep}{0.8em} 
{\renewcommand{\arraystretch}{1.3}
\begin{tabular}{l|ccc}
\hline
\textbf{Dataset}   & \textbf{T2D} & \textbf{Limaye} & \textbf{Wikipedia}    \\ \hline
\textbf{T2K}       & 0.97         & 1.00      & 0.8899    \\
\textbf{TabEAno}    & 1.00         & 1.00      & 0.8987     \\ \hline
\end{tabular}}
\end{table}

\subsubsection{Entity Annotation}
Table \ref{tab:res_entity} depicts the results of entity annotation by Precision (P), Recall (R), and F1-score (F1) on the T2D, Limaye, and Wikipedia tables dataset.

\begin{table}[!ht]
\centering
\caption{Results of the entity annotation by Precision (P), Recall (R), and F1 score (F1) on the T2D, Limaye, and Wikipedia tables dataset. The best score when running on 100\% dataset is in \textbf{bold}. The results of this table are adapted from Ritze et al.\cite{ritze2015matching}, Efthymiou et al. \cite{efthymiou2017matching}, and Zhang et al. \cite{www2020}}
\label{tab:res_entity}
\setlength{\tabcolsep}{0.4em} 
{\renewcommand{\arraystretch}{1.5}
\begin{threeparttable}
\begin{tabular}{l|lll|lll|ccc}
\hline
\multicolumn{1}{c|}{\multirow{2}{*}{\textbf{Method}}} & \multicolumn{3}{c|}{\textbf{T2D}}                                                & \multicolumn{3}{c|}{\textbf{Limaye}}                                             & \multicolumn{3}{c}{\textbf{Wikipedia}}                                                     \\ \cline{2-10} 
\multicolumn{1}{c|}{}                        & \multicolumn{1}{c}{P} & \multicolumn{1}{c}{R} & \multicolumn{1}{c|}{F1} & \multicolumn{1}{c}{P} & \multicolumn{1}{c}{R} & \multicolumn{1}{c|}{F1} & P                         & R                         & F1                         \\ \hline
String Matching\tnote{*}                              & 0.53                 & 0.53                 & 0.53                   & \multicolumn{1}{c}{-} & \multicolumn{1}{c}{-} & \multicolumn{1}{c|}{-}  & -                         & -                         & -                          \\
DBpedia Lookup\tnote{*}                                 & 0.79                 & 0.73                 & 0.76                   & \multicolumn{1}{c}{-} & \multicolumn{1}{c}{-} & \multicolumn{1}{c|}{-}  & -                         & -                         & -                          \\
T2K\cite{ritze2015matching}\tnote{*}                                           & 0.90                 & 0.76                 & 0.82                   & \multicolumn{1}{c}{-} & \multicolumn{1}{c}{-} & \multicolumn{1}{c|}{-}  & -                         & -                         & -                          \\ \hline
Le\&Titov\cite{le_titov}\tnote{\textdagger}                                     & 0.91                & 0.83                 & 0.87                   & 0.81                 & 0.83                 & 0.82                   & -                         & -                         & -                          \\
Wikipedia Search\cite{www2020}\tnote{\textdagger}                              & 0.91                 & 0.82                 & 0.86                   & 0.77                 & 0.82                 & 0.80                   & -                         & -                         & -                          \\
Wikipedia Search+ED\cite{www2020}\tnote{\textdagger}                          & 0.91                 & 0.85                 & 0.88                   & 0.77                 & 0.81                 & 0.79                   & -                         & -                         & -                          \\
Rerank+ED\cite{www2020}\tnote{\textdagger}                                     & 0.94                 & 0.86                 & 0.90                   & 0.96                 & 0.89                 & 0.93                   & -                         & -                         & -                          \\ \hline \hline
T2K\cite{ritze2015matching}\tnote{\textdaggerdbl}                                           & \multicolumn{1}{c}{-} & \multicolumn{1}{c}{-} & \multicolumn{1}{c|}{-}  & 0.70                 & 0.63                 & 0.66                   & \multicolumn{1}{l}{0.70} & \multicolumn{1}{l}{0.22} & \multicolumn{1}{l}{0.34} \\
DBpedia Refined\cite{efthymiou2017matching}\tnote{\textdaggerdbl}                              & 0.86                 & 0.76                 & 0.81                   & 0.73                 & 0.68                 & 0.71                   & -                         & -                         & -                          \\
FactBase Lookup\cite{efthymiou2017matching}\tnote{\textdaggerdbl}                               & 0.88                 & 0.78                 & 0.83                   & 0.84                 & 0.78                 & 0.81                   & \multicolumn{1}{l}{0.70} & \multicolumn{1}{l}{0.50} & \multicolumn{1}{l}{0.58} \\
Embeddings\cite{efthymiou2017matching}\tnote{\textdaggerdbl}                                    & 0.86                 & 0.77                 & 0.81                   & \multicolumn{1}{l}{0.84} & \multicolumn{1}{l}{0.65} & \multicolumn{1}{l|}{0.73}  & \multicolumn{1}{l}{0.70} & \multicolumn{1}{l}{0.53} & \multicolumn{1}{l}{0.60} \\
Blocking\cite{blocking}\tnote{\textdaggerdbl}                                      & 0.32                 & 0.71                 & 0.44                   & \multicolumn{1}{c}{-} & \multicolumn{1}{c}{-} & \multicolumn{1}{c|}{-}  & \multicolumn{1}{l}{0.16} & \multicolumn{1}{l}{0.39} & \multicolumn{1}{l}{0.23} \\
LogMap\cite{logmap}\tnote{\textdaggerdbl}                                        & 0.89                 & 0.57                 & 0.70                   & \multicolumn{1}{c}{-} & \multicolumn{1}{c}{-} & \multicolumn{1}{c|}{-}  & \multicolumn{1}{l}{0.34} & \multicolumn{1}{l}{0.29} & \multicolumn{1}{l}{0.32} \\
PARIS\cite{paris}\tnote{\textdaggerdbl}                                          & 0.42                 & 0.04                 & 0.07                   & \multicolumn{1}{c}{-} & \multicolumn{1}{c}{-} & \multicolumn{1}{c|}{-}  & -                         & -                         & -                          \\
Hybrid I\cite{efthymiou2017matching}\tnote{\textdaggerdbl}                                       & 0.87                 & 0.83                 & 0.85                   & 0.84                 & 0.79                 & 0.81                   & \multicolumn{1}{l}{0.66} & \multicolumn{1}{l}{0.57} & \multicolumn{1}{l}{0.61} \\
Hybrid II\cite{efthymiou2017matching}\tnote{\textdaggerdbl}                                      & 0.85                 & 0.81                 & 0.83                   & 0.84                 & 0.79                 & 0.82                   & \multicolumn{1}{l}{\textbf{0.69}} & \multicolumn{1}{l}{0.60} & \multicolumn{1}{l}{0.64} \\ \hline
TabEAno                                       & \textbf{0.92}                 & \textbf{0.90}                 & \textbf{0.91}                   & \textbf{0.89}                 & \textbf{0.87}                 & \textbf{0.88}                   & \multicolumn{1}{l}{0.63} & \multicolumn{1}{l}{\textbf{0.67}} & \multicolumn{1}{l}{\textbf{0.65}} \\ \hline
\end{tabular}
\begin{tablenotes}
        \footnotesize
        \item[*] Results were reported by Ritze et al.\cite{ritze2015matching}, these were evaluated on 50\% of the T2D dataset, the remaining 50\% data is the optimize set. The reported F1 of T2K on the optimized set is 0.86. 
        \item[\textdagger] Results are taken from Zhang et al.\cite{www2020}. The results were conducted on 20\% of T2D and Limaye dataset, 80\% of data is used as training data.
        \item[\textdaggerdbl] Results are taken from Efthymiou et al. \cite{efthymiou2017matching} on 100\% T2D, Limaye, and Wikipedia tables dataset.     
    \end{tablenotes}
\end{threeparttable}
}
\end{table}

The three first rows of the results are taken from Ritze et al. \cite{ritze2015matching}. These results were evaluated on 50\% of the T2D dataset, the remaining 50\% T2D data was used as an optimizing set to find the best parameters for T2K. The T2K results on 100\% Limaye, and Wikipedia tables dataset were reported by Efthymiou et al. \cite{efthymiou2017matching} (the 8th row). 

The results from the 4th row to the 7th row were reported by \cite{www2020}. These methods were tested on 20\% of T2D, and Limaye dataset, while 80\% of data were used as training data. Note that, besides the table as the input, Zhang et al. also assume the information about the core attribute and table header available as input data. Le\&Titov \cite{le_titov} is an entity linking for documents based on deep learning. The F1-score of Wikipedia Search \cite{www2020} achieves higher performance than FactBase Lookup, Hybrid I, and II. Despised the different environment settings, it is also verified our observation and Efthymiou et al. \cite{efthymiou2017matching} that perform a lookup on the latest knowledge could get better performance than the previous one (T2D was matched to DBpedia 2014, and Limaye was matched to DBpedia 2015-10). Using Entity Disambiguation (ED) as a classifier also improves the performance of entity annotation. The result of Rerank+ED achieves the best performance as the report of Zhang et al. \cite{www2020} also validates that performing training on annotated data might improve the performance of entity annotation.

The results from the 8th row to the 16th row were reported by Efthymiou et al. \cite{efthymiou2017matching}. These methods were tested on 100\% of T2D, Limaye, and Wikipedia tables datasets. DBpedia Refined use DBpedia lookup results, while FactBase Lookup aggregate many resources as Wikipedia, DBpedia, FreeBase, YAGO, and both methods have a re-find step to select the best entity candidates with the constraint of entity types, and entity attributes. The Embeddings approach is an global disambiguation inspired from DoSeR framework \cite{DBLP:conf/esws/ZwicklbauerSG16DoSeR}. The Blocking\cite{blocking}, LogMap \cite{logmap}, and PARIS \cite{paris} are these ontology matching frameworks. The Hybrid approaches are the ensemble system that combines the annotation results of FactBase Lookup and Embeddings. Hybrid I select the result of FactBase first if there is no annotation, the system uses the results of Embeddings, while Hybrid II is reverted process as using the result from Embbedings first and then the result of FactBase after.  

Because of the difference in evaluation setting with Ritze et al. \cite{ritze2015matching} (evaluated on 50\% of T2D data), and Zhang et al. \cite{www2020} (evaluated on 20\% of T2D, and Limaye dataset), we only compare the results of TabEAno with the annotation results reported by Efthymious et al. \cite{efthymiou2017matching} (evaluated on 100\% of T2D, Limaye, and Wikipedia tables datasets) . Overall, TabEAno outperforms the-state-of-the-art (SOTA) Hybrid I and II. TabEAno results outperform Hybrid I with a difference of +0.06 F1 score in T2D, and Hybrid II with the same difference as +0.06 F1 score in Limaye dataset. In the Wikipedia tables dataset, the improvement is 0.01 in F1-score but TabEAno provides a higher number of annotations since it got +0.07 in Recall metrics. Although TabEAno does not need the training process, it can achieve the better results in T2D dataset than Rerank+ED\cite{www2020}. 

\section{Conclusion} \label{sec:conclusion}
In this paper, we propose TabEAno, an unsupervised approach to automatically annotate table rows with knowledge graph entities. We propose an extension heuristic for header, and core attribute detection which performs effectively in the three tested datasets. We also introduce a novel ``two-cells" lookup method based on the assumption that there is a logical relation (the knowledge graph edge) between the two closed cells in a table. Despite the simplicity of the approach, the TabEAno outperforms the state of the art approaches (Hybrid I and II) \cite{efthymiou2017matching} with a difference of +0.06 in the F1 score in T2D and Limaye data. The evaluation in the large-scale setting also shows that TabEAno outperforms the state of the art approaches HyBrid II\cite{efthymiou2017matching} with an increase of +0.01 in terms of the F1 score. Moreover, compare with other systems, TabEAno also achieves better recall in general, as the ability to make more annotations. 

In future work, TabEAno could be improved by relaxing its assumptions as follows:
\begin{itemize}
    \item Closed-world assumption: Using Embeddings \cite{efthymiou2017matching} might help to improve the performance of entity annotations in cases the long-tail entities, or missing the corresponding attributes. The pre-trained embedding from contextualized words embedding on tables also is an interesting direction \cite{DBLP:journals/corr/abs-2004-02349tapas, DBLP:journals/corr/abs-2005-08314tabert}. 
    \item Relational vertical table type assumption: The ``two cells" lookup of TabEAno could be extended to other table types. Instead of performing a query on two closed cells in the same row as in the relational vertical types, we can query on two any cells in the same column, row, or neighbor cells. Then, we can re-find the table structure by using the information about number entity candidates of lookup results and table column, or row majority voting. Despite the high cost of query combinations, it is an interesting direction. 
    \item Assumption of the input tables are independent: In practice, many tables could have shared schema. For example, tables on the same domain in the Web, or tables in data resources with a different version of Open Data Portals. Stitching those tables also improves the performance of entity annotation \cite{Lehmberg:2017:SWT:3137628.3137657}. 
\end{itemize}

%
%
%
\bibliographystyle{splncs04}
\bibliography{mybibliography}

\begin{thebibliography}{10}
\providecommand{\url}[1]{\texttt{#1}}
\providecommand{\urlprefix}{URL }
\providecommand{\doi}[1]{https://doi.org/#1}

\bibitem{ahmadov2015towards}
Ahmadov, A., Thiele, M., Eberius, J., Lehner, W., Wrembel, R.: Towards a hybrid
  imputation approach using web tables. In: 2015 IEEE/ACM 2nd International
  Symposium on Big Data Computing (BDC). pp. 21--30 (2015)

\bibitem{tabel}
Bhagavatula, C.S., Noraset, T., Downey, D.: Tabel: Entity linking in web
  tables. In: {ISWC} 2015. Lecture Notes in Computer Science, vol.~9366, pp.
  425--441. Springer (2015)

\bibitem{colnet}
Chen, J., Jim{\'{e}}nez{-}Ruiz, E., Horrocks, I., Sutton, C.A.: Colnet:
  Embedding the semantics of web tables for column type prediction. In: {AAAI}
  2019,. pp. 29--36. {AAAI} Press (2019)

\bibitem{blocking}
Dalvi, B.B., Cohen, W.W., Callan, J.: Websets: extracting sets of entities from
  the web using unsupervised information extraction pp. 243--252 (2012)

\bibitem{Dong2014}
Dong, X., Gabrilovich, E., Heitz, G., Horn, W., Lao, N., Murphy, K., Strohmann,
  T., Sun, S., Zhang, W.: Knowledge vault: A web-scale approach to
  probabilistic knowledge fusion. In: Proceedings of the 20th ACM SIGKDD
  international conference on Knowledge discovery and data mining. pp.
  601--610. ACM (2014)

\bibitem{efthymiou2017matching}
Efthymiou, V., Hassanzadeh, O., Rodriguez{-}Muro, M., Christophides, V.:
  Matching web tables with knowledge base entities: From entity lookups to
  entity embeddings. In: ISWC. pp. 260--277 (2017)

\bibitem{DBLP:conf/www/ErlingM09}
Erling, O., Mikhailov, I.: Faceted views over large-scale linked data. In:
  {WWW2009}, {LDOW} 2009. {CEUR} Workshop Proceedings, vol.~538. CEUR-WS.org
  (2009)

\bibitem{hassanzadeh2012record}
Hassanzadeh, O.: Record linkage for web data. Ph.D. thesis (2012)

\bibitem{DBLP:journals/corr/abs-2004-02349tapas}
Herzig, J., Nowak, P.K., M{\"{u}}ller, T., Piccinno, F., Eisenschlos, J.M.:
  {TAPAS:} weakly supervised table parsing via pre-training. CoRR
  \textbf{abs/2004.02349} (2020)

\bibitem{spacy2}
Honnibal, M., Montani, I.: {spaCy 2}: Natural language understanding with
  {B}loom embeddings, convolutional neural networks and incremental parsing
  (2017), to appear

\bibitem{logmap}
Jim{\'{e}}nez{-}Ruiz, E., Grau, B.C.: Logmap: Logic-based and scalable ontology
  matching. In: ISWC. vol.~7031, pp. 273--288. Springer (2011)

\bibitem{DBLP:conf/semweb/2019semtab}
Jim{\'{e}}nez{-}Ruiz, E., Hassanzadeh, O., Srinivas, K., Efthymiou, V., Chen,
  J. (eds.): Proceedings of the Semantic Web Challenge on Tabular Data to
  Knowledge Graph Matching co-located with the 18th International Semantic Web
  Conference, SemTab@ISWC 2019, Auckland, New Zealand, October 30, 2019, {CEUR}
  Workshop Proceedings, vol.~2553. CEUR-WS.org (2020)

\bibitem{le_titov}
Le, P., Titov, I.: Improving entity linking by modeling latent relations
  between mentions. In: Proceedings of {ACL} 2018, Volume 1: Long Papers. pp.
  1595--1604. Association for Computational Linguistics (2018)

\bibitem{Lehmberg:2017:SWT:3137628.3137657}
Lehmberg, O., Bizer, C.: Stitching web tables for improving matching quality.
  Proc. VLDB Endow.  \textbf{10}(11),  1502--1513 (Aug 2017)

\bibitem{Lehmberg2016}
Lehmberg, O., Ritze, D., Meusel, R., Bizer, C.: A large public corpus of web
  tables containing time and context metadata. In: Proceedings of the 25th
  International Conference Companion on World Wide Web. pp. 75--76.
  International World Wide Web Conferences Steering Committee (2016)

\bibitem{Lehmberg2015}
Lehmberg, O., Ritze, D., Ristoski, P., Meusel, R., Paulheim, H., Bizer, C.: The
  mannheim search join engine. Web Semantics: Science, Services and Agents on
  the World Wide Web  \textbf{35}(P3),  159--166 (2015)

\bibitem{DBLP:journals/pvldb/LimayeSC10}
Limaye, G., Sarawagi, S., Chakrabarti, S.: Annotating and searching web tables
  using entities, types and relationships. {PVLDB}  \textbf{3}(1),  1338--1347
  (2010)

\bibitem{Mitlohner2016}
Mitlöhner, J., Neumaier, S., Umbrich, J., Polleres, A.: Characteristics of
  open data csv files. In: 2016 2nd International Conference on Open and Big
  Data (OBD). pp. 72--79 (2016)

\bibitem{Nargesian2018}
Nargesian, F., Zhu, E., Pu, K.Q., Miller, R.J.: Table union search on open
  data. Proceedings of the VLDB Endowment  \textbf{11}(7),  813--825 (2018)

\bibitem{mtab}
Nguyen, P., Kertkeidkachorn, N., Ichise, R., Takeda, H.: Mtab: Matching tabular
  data to knowledge graph using probability models. In: SemTab@ISWC 2019.
  {CEUR} Workshop Proceedings, vol.~2553, pp. 7--14. CEUR-WS.org (2019)

\bibitem{embnumplus}
Nguyen, P., Nguyen, K., Ichise, R., Takeda, H.: Embnum+: Effective, efficient,
  and robust semantic labeling for numerical values. New Gener. Comput.
  \textbf{37}(4),  393--427 (2019)

\bibitem{Nguyen2015}
Nguyen, T.T., Nguyen, Q.V.H., Weidlich, M., Aberer, K.: Result selection and
  summarization for web table search. In: Data Engineering (ICDE), 2015 IEEE
  31st International Conference on. pp. 231--242. IEEE (2015)

\bibitem{Nishida2017}
Nishida, K., Sadamitsu, K., Higashinaka, R., Matsuo, Y.: Understanding the
  semantic structures of tables with a hybrid deep neural network architecture.
  In: AAAI. pp. 168--174 (2017)

\bibitem{pham16:iswc}
Pham, M., Alse, S., Knoblock, C.A., Szekely, P.: Semantic labeling: a
  domain-independent approach. In: International Semantic Web Conference. pp.
  446--462. Springer (2016)

\bibitem{ritze2015matching}
Ritze, D., Lehmberg, O., Bizer, C.: Matching html tables to dbpedia. In: WIMS.
  p.~10. ACM (2015)

\bibitem{Sekhavat2014}
Sekhavat, Y.A., Di~Paolo, F., Barbosa, D., Merialdo, P.: Knowledge base
  augmentation using tabular data. In: LDOW (2014)

\bibitem{speer-2019-ftfy}
Speer, R.: ftfy. Zenodo (2019), version 5.5

\bibitem{paris}
Suchanek, F.M., Abiteboul, S., Senellart, P.: {PARIS:} probabilistic alignment
  of relations, instances, and schema. Proc. {VLDB} Endow.  \textbf{5}(3),
  157--168 (2011)

\bibitem{DBLP:journals/corr/abs-2005-08314tabert}
Yin, P., Neubig, G., Yih, W., Riedel, S.: Tabert: Pretraining for joint
  understanding of textual and tabular data. CoRR  \textbf{abs/2005.08314}
  (2020)

\bibitem{Zhang2013}
Zhang, M., Chakrabarti, K.: Infogather+: semantic matching and annotation of
  numeric and time-varying attributes in web tables. In: ICMD. pp. 145--156.
  ACM (2013)

\bibitem{www2020}
Zhang, S., Meij, E., Balog, K., Reinanda, R.: Novel entity discovery from web
  tables. In: {WWW} '20. pp. 1298--1308. {ACM} / {IW3C2} (2020)

\bibitem{zhang2017effective}
Zhang, Z.: Effective and efficient semantic table interpretation using
  tableminer+. Semantic Web  \textbf{8}(6),  921--957 (2017)

\bibitem{DBLP:conf/esws/ZwicklbauerSG16DoSeR}
Zwicklbauer, S., Seifert, C., Granitzer, M.: Doser - {A}
  knowledge-base-agnostic framework for entity disambiguation using semantic
  embeddings. In: {ESWC} 2016. Lecture Notes in Computer Science, vol.~9678,
  pp. 182--198. Springer (2016)

\end{thebibliography}
\end{document}